\documentclass[12pt]{l4dc2021alt}

\title[Tracking and Planning with Spatial World Models]{Tracking and Planning with Spatial World Models}
\usepackage{times}
\usepackage{todonotes}
\usepackage{wrapfig}
\usepackage{cleveref}
\usepackage{supertabular}
\usepackage{multicol}
\usepackage{multirow}
\usepackage{ulem}
\usepackage{xcolor}
\usepackage{booktabs}
\usepackage[export]{adjustbox}
\usepackage{appendix}
\usepackage{mymath}

\newcommand\myeq{\mkern1.5mu{=}\mkern3.8mu}

\author{\Name{Bar\i \c{s} Kayal\i bay}\thanks{Equal contribution.} \Email{bkayalibay@argmax.ai}\\
	\Name{Atanas Mirchev}\footnotemark[1] \Email{atanas.mirchev@argmax.ai}\\
 \Name{Patrick {van der Smagt}}\\
 \Name{Justin Bayer} \Email{bayerj@argmax.ai}\\
 \addr Machine Learning Research Lab, Volkswagen Group, Munich}

\begin{document}

\maketitle

\begin{abstract}%
We introduce a method for real-time navigation and tracking with differentiably rendered world models.
Learning models for control has led to impressive results in robotics and computer games, but this success has yet to be extended to vision-based navigation.
To address this, we transfer advances in the emergent field of differentiable rendering to model-based control.
We do this by planning in a learned 3D spatial world model, combined with a pose estimation algorithm previously used in the context of TSDF fusion, but now tailored to our setting and improved to incorporate agent dynamics.
We evaluate over six simulated environments based on complex human-designed floor plans and provide quantitative results.
We achieve up to 92\% navigation success rate at a frequency of 15 Hz using only image and depth observations under stochastic, continuous dynamics.
\end{abstract}
 

\section{Introduction}

\begin{wrapfigure}{r}{0.5\textwidth}
  \vspace{-1.5em}
  \begin{center}
      \includegraphics[width=0.95\linewidth]{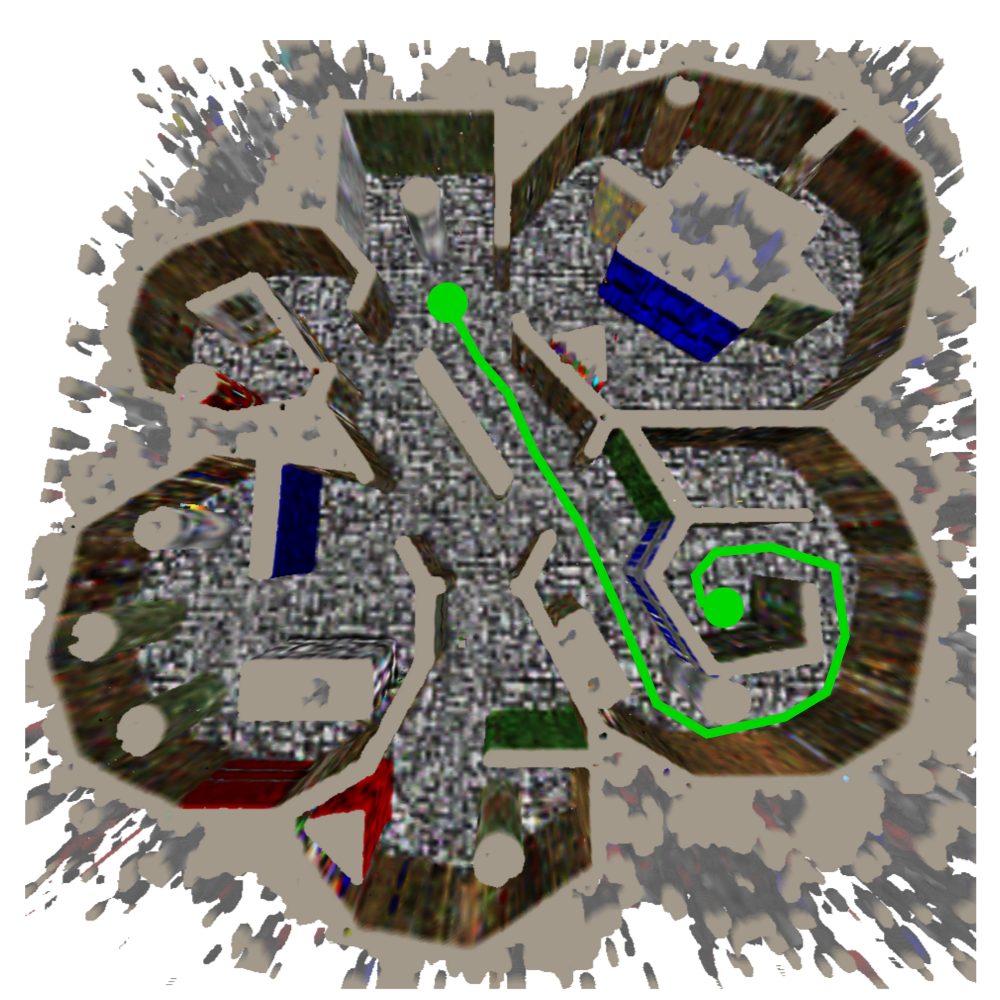}
  \end{center}
  \caption{
      Planning in a world model. Isometric model view with planned trajectory.
  }
  \vspace{-4.0em}
\end{wrapfigure}
Autonomous systems need to understand their environment to make good decisions.
Optimal control and reinforcement learning therefore hinge on learning or engineering accurate models of both dynamics and observations, which allow us to plan into the future and find optimal actions.
The paradigm of learning models has been successfully applied to Atari games \citep[e.g.][]{kaiser2020modelbased}, walking with complex simulated robots \citep[e.g.][]{Hafner2019learning}, controlling unmanned aerial vehicles \citep[e.g.][]{becker2020learning} and meta-reinforcement learning for robotics \citep[e.g.][]{zhao2020meld}.

Prior work has attempted to learn such world models of spatial environments \citep{fraccaro2018generative, mirchev2019approximate}.
These early models were limited to simple scenes, which restricted their practical use in control to toy scenarios \citep{kayalibay2018navigation}.
Their limitations were a result of attempting to model the complex interaction of a camera or depth sensor with its environment using a combination of neural networks and simple inductive biases.
More recently, this balance has shifted in favour of the inductive bias, with an increased focus on engineering and domain knowledge \citep{mildenhall2020nerf,mirchev2021variational,sitzmann2020scene,mueller2019neural}.
The result is the field of differentiable rendering, which models camera sensing by applying multiple-view geometry and adapting concepts from computer graphics, with neural networks being relegated to the relatively simple task of predicting properties like colour or occupancy over 3D space.

Differentiable rendering allows to model spatial environments with outstanding fidelity.
It is then natural to use such models for control, an idea that was voiced as early as the 1990s \citep{ore97mobile}.
This has been done by \cite{li20213d} in the context of multi-link robots and \cite{adamkiewicz21vision} in the context of navigation, albeit without a quantitative evaluation of success.

Here we present a practical navigation algorithm for planar robots which uses a learned world model.
Navigating based on visual observations alone requires inferring the agent's pose.
We will therefore address the problem of state estimation as well.
Our exact contributions are:

\begin{itemize}
    \item We present a 15-Hz real-time algorithm for navigation under noisy agent dynamics and only local RGB-D observations in world models obtained with differentiable rendering.
    \item We test our method on simulated yet challenging indoor environments based on actual floor plans and provide a thorough quantitative evaluation.
    \item We improve an existing method for tracking camera poses by using the dynamics of the agent and apply it to our setting.
          This runs about five times faster than propagating gradients through the renderer, and is thus well-suited for vision-based control.
\end{itemize}
\section{Related Work}

\noindent\textbf{Generative models and maps.} Various generative models of space have been proposed \citep{fraccaro2018generative, planche2019incremental, gregor2019shaping, mirchev2019approximate}.
Among these, the work of \cite{mirchev2019approximate} was later adapted to a navigation task with the addition of $\text{A}^*$-based planning \citep{kayalibay2018navigation}.
The method was demonstrated on simple 2D mazes with $360^\circ$ depth range observations.
In contrast, others have proposed models aimed at tasks like navigation and exploration, which do not require generating new observations from the model.
These approaches define read and write operations for distilling information from camera images into a 2D top-down map of the environment.
Notable contributions here include the work of \citet{parisotto2017neural}, where a 2D neural map with abstract features is written to and read from to solve searching problems.
\cite{gupta2019cognitive} proposed a similar algorithm, extended by a planning module for navigation.
The write operation was improved using projective geometry in \citep{chen2019learning} and \citep{chaplot2020learning}, the latter achieving impressive navigation performance.
\cite{ramakrishnan2020occupancy} extended the setup by learning to predict occupancy beyond visible locations.

\noindent\textbf{Differentiable rendering, pose estimation and control.} Various methods for differentiably rendering images have been proposed in recent years.
Of these, the method of \cite{mildenhall2020nerf} relies on modeling colour and spatial density with neural nets.
\cite{sitzmann2020scene} used an LSTM to find the point where a ray would hit the scene geometry.
\cite{lombardi2019neural} proposed using voxel grids and warping fields to encode and decode scenes.
\cite{mirchev2021variational} employ voxel grids as well, focusing on localisation and mapping with unmanned aerial vehicles (UAV).
Taking a different route, \cite{niemeyer2020differentiable} and \cite{yariv2020multiview} rely on implicit differentiation to define the surface of the scene.

Several works deal with pose estimation in the context of differentiable rendering.
\cite{wang2021nerf} jointly optimise pose, camera and scene parameters in a set of simple scenes.
Assuming that the scene parameters have been learned in advance, \cite{yenchen2021inerf} propose finding camera poses using gradient descent and an efficient pixel subsampling scheme.
Notably, \cite{sucar2021imap} propose a real-time algorithm for joint tracking and mapping, though their evaluation is limited to small scenes.
\cite{mirchev2021variational} introduce a SLAM algorithm based on differentiable rendering, capable of tracking a UAV in photorealistically rendered simulations, though not in real time.

To the best of our knowledge, only two papers exist that apply differentiable rendering in a control context.
Of these, \citep{li20213d} focuses on manipulation tasks involving a robot arm.
\cite{adamkiewicz21vision} investigate using neural radiance fields to solve navigation tasks.
Their work is close to ours, though we differ in a couple of points.
We focus on navigation with planar robots in indoor environments, and provide a quantitative evaluation on a set of complex scenes based on human-designed floor plans which were converted into levels for the Vizdoom simulator \citep{wydmuch2018vizdoom}.
In contrast, they provide experiments with different kinds of robots, though they do not report any quantitative results indicating large-scale navigation success.
Our work also approaches the problem of pose estimation differently.
They approach pose estimation by combining the approach of \cite{yenchen2021inerf} with a transition loss and obtain a Bayesian filter.
The benefit of our approach is its speed over computing gradients through the rendering pipeline.
Finally, they propose a gradient-based algorithm for obstacle avoidance, which is initialised using $\text{A}^*$-search.
We similarly use $\text{A}^*$-search, and track the planned trajectory using a simple low-level controller intended for a planar robot.

\section{Background}
\subsection{Differentiable Rendering}
The backbone of most differentiable rendering approaches is a parametric function that maps 3D space to geometric information (e.g. occupancy) and colour.
Formally, we have two maps $f: \mathbb{R}^3 \rightarrow \mathbb{R}$ and $g: \mathbb{R}^3 \rightarrow \mathbb{R}^3$, where one learns the occupancy (alternatively, density or opacity) and the other corresponds to the RGB value.
The colour model might also take a viewing angle as input, which allows modeling reflections.
We can then generate an RGB-D image from any camera pose in the special Euclidean group $\mathrm{SE}(3)$ by defining a differentiable rendering procedure.

The first step of the rendering procedure involves finding the point where a ray intersects the scene geometry for the first time, the latter being implicitly defined by the occupancy model.
We follow the approach of \cite{mirchev2021variational}, where the occupancy function is linearised around the point where it first exceeds a threshold, and the intersection point is then found using that linear approximation.
This way of rendering is well-suited to our purposes, since it directly defines the occupancy of a point in space, which defines where the obstacles are when navigating.
Formally, given a ray parameterised by the camera centre and an offset vector: $\mathbf{r} = \mathbf{c} + k\mathbf{v}$, we evaluate the occupancy function for $k \in \{\Delta, 2\Delta, \dots, n\Delta\}$, where $\Delta$ gives the distance between two points along the ray, and $n\Delta$ is the maximum range of the camera.
Then, if $\mathbf{p}^+ = \mathbf{c} + k^+\mathbf{v}$ is the first point for which the occupancy function returns a value above some threshold $\tau$, and $\mathbf{p}^- = \mathbf{c} + k^-\mathbf{v}$ is the point that precedes it, the intersection point $\mathbf{p}$ is defined as $\mathbf{p}^* = \mathbf{c} + k^*\mathbf{v}$ for:
\begin{equation}
	k^* = \alpha k^+ + (1 - \alpha)k^-,\quad \alpha = \dfrac{\tau - f(p^-)}{f(p^+) - f(p^-)}.
\end{equation}                                                          
									  
The setup of Mirchev et al.\ (2021) is also set apart from other differentiable rendering approaches as voxel grids are used instead of neural nets to capture the scene.
Making use of trilinear interpolation, differentiability is maintained and hence the use of gradient-based techniques possible.
We follow this decision as well, since indexing a voxel grid via interpolation is faster than evaluating a neural net.
This reduces the time it takes to render a full image, a step required by our tracking method (we discuss rendering speed in \cref{sec:experiments}).
We note that the voxel grid can be seen as an ensemble of small neural networks with partially fixed weights, where the networks are distributed over space in a grid-fashion, resembling the work of \cite{reiser2021kilonerf}.

Formally the occupancy map is a 3D tensor $\mathcal{M}^{\text{occ}}\in \mathbb{R}^{h\times w\times d}$ and the colour map is a 4D tensor $\mathcal{M}^{\text{col}}\in\mathbb{R}^{h\times w\times d\times3}$.
The occupancy and colour functions $f$ and $g$ are then given by extracting the eight neighbor cells of the input point and trilinearly interpolating between them.
The density of an RGB-D image $\mathbf{x}$ taken from the camera pose $\mathbf{z}$ under the differentiably rendered model is then:
\begin{equation}
	p(\mathbf{x} \mid \mathbf{z}) = \prod_{i, j}  p(\mathbf{x}_{ij} \mid \mathbf{z}, \mathcal{M}^{\text{occ}}, \mathcal{M}^{\text{col}}) = \prod_{\mathbf{x}_{ij} \in \mathbf{x}} \text{Laplace} \left(\begin{bmatrix} \mathbf{x}_{ij}^{\text{rgb}} \\ \mathbf{x}_{ij}^\text{d} \end{bmatrix} \middle\vert \begin{bmatrix} g(\mathbf{p}^*) \\ k^* \end{bmatrix}, \begin{bmatrix} \sigma_1 \\ \sigma_2 \end{bmatrix} \right).
\end{equation}
Here, $\mathbf{x}_{ij}$ is a pixel from the image, and $\mathbf{x}_{ij}^{\text{rgb}}$ and $\mathbf{x}_{ij}^\text{d}$ are its RGB and depth readings respectively.
$\mathbf{p}^*$ and $k^*$ are the intersection point of the ray and its distance from the camera.
The Laplace distribution's scales for colour and depth are parameterised by $\sigma_1$ and $\sigma_2$, which are treated as learnable parameters.
The ray for a pixel is given by: $\mathbf{T}(\mathbf{z}) + d\mathbf{R}(\mathbf{z})\mathbf{K}^{-1}\begin{bmatrix} i & j & 1\end{bmatrix}^T$, with $\mathbf{T}(\cdot)$ and $\mathbf{R}(\cdot)$ translation and rotation defined by the camera pose, $\mathbf{K}$ the camera intrinsic matrix, $(i, j)$ the 2D coordinates of the pixel and $d$ a depth reading.

This differentiable renderer was proposed by \cite{mirchev2021variational} in the context of online simultaneous localisation and mapping.
Here the map is learned offline on a set of RGB-D images with known camera poses.
Given a dataset of images and camera poses $\mathcal{D} = \{(\mathbf{x}^{(i)}, \mathbf{z}^{(i)})\}_{i=1}^N$, the occupancy and colour parameters $\mathcal{M}^{\text{occ}}$ and $\mathcal{M}^{\text{col}}$ are obtained by gradient-based minimisation of:
\begin{equation}
	\mathcal{L}(\mathcal{M}^{\text{occ}}, \mathcal{M}^{\text{col}}) = -\sum_{\mathbf{x}, \mathbf{z} \in \mathcal{D}} \sum_{i, j} \log p(\mathbf{x}_{ij} \mid \mathbf{z}, \mathcal{M}^{\text{occ}}, \mathcal{M}^{\text{col}}).
	\label{eq:obj}
\end{equation}
We approximate the gradients by Monte-Carlo sampling images and pixels uniformly for speed.

\subsection{Pose Estimation with Differentiable Renderers} \label{sec:pointtoplane}
Agents moving about freely normally cannot directly observe their state.
Instead, they need to rely on estimating it from observations.
Pose estimation in prior approaches with differentiable rendering has so far focused on backpropagating pixel-wise reconstruction errors through the renderer \citep{yenchen2021inerf,sucar2021imap,adamkiewicz21vision}:
\eq{
    \arg\min_{\state}\, -\log p(\obs \mid \state, \map\occsup, \map\colsup).
}
We additionally consider a different route for state estimation, optimising both photometric and point-to-plane reprojection errors~\citep{chen1992object}.
Given an RGB-D image $\hat \obs = (\,\hat\obs^\rgb, \hat\obs^\depth)$ rendered by the model, a new observation $\obs = (\,\obs^\rgb, \obs^\depth)$ and a relative pose $\state$ we minimise:
\eq{
    \arg\min_{\state} \sum_k \norm{\hat\obs^\rgb[\pi(\transform_\state\point^k)] - \obs^\rgb[\pi(\point^k)]}_1 + \sum_k \norm{\langle \hat\point^k- \transform_\state\point^k, \hat\normal^k \rangle}_1. \numberthis \label{eq:reproject}
}
Here, $\point^k$ is a 3D point in the observed camera frame\footnote{In practice, we obtain $\point^k$ from pixel coordinates, depth $d^k$ and the inverse intrinsic matrix $\mathbf{K}^{-1}$ in the observed image. We use it directly in \cref{eq:reproject} for brevity.} and $\hat \point^k$ is a 3D point in the camera frame of the model prediction, related via projective association \citep{blais1995registering,stotko2016state}.
$\pi$~denotes perspective projection from 3D into the image plane, using the known intrinsic matrix $\mathbf{K}$.
The dot product $\langle\,\cdot\,,\hat\normal^k\rangle$ with the corresponding normal vector $\hat \normal^k$ defines the point-to-plane objective.
First, normals are computed based on image gradients of the rendered prediction $\hat\obs^\depth$.
Then the difference $\hat\point^k- \transform_\state\point^k$ is projected onto the normal of the rendered surface, therefore reflecting the distance between the transformed point $\mathbf{T}\point^k$ and a hyperplane tangent to the surface.
It drives points from the second camera frame to align to the surface implicitly defined by the rendered depth $\hat\obs^\depth$.
The L1-norm (Laplace assumption) is used for increased robustness to outliers.

This method is commonly referred to as point-to-plane ICP \citep{chen1992object} with photometric constraints \citep{steinbrucker2011real,audras2011real}.
Variants of this approach have been used to good effect for tracking new RGB-D observations w.r.t.\ a fused scene model \citep{newcombe2011kinectfusion,niessner2013real,whelan2015elasticfusion}.
Traditionally, this is well-explored for maps obtained via TSDF fusion.
We show the same concept is applicable to maps obtained by optimisation through differentiable rendering.
In \cref{stateestimation} we will discuss its advantages compared to gradient descent directly through the renderer.
\section{Method}

\subsection{Motion Planning}

\begin{figure}[t]
            \centering
            \includegraphics[width=\linewidth]{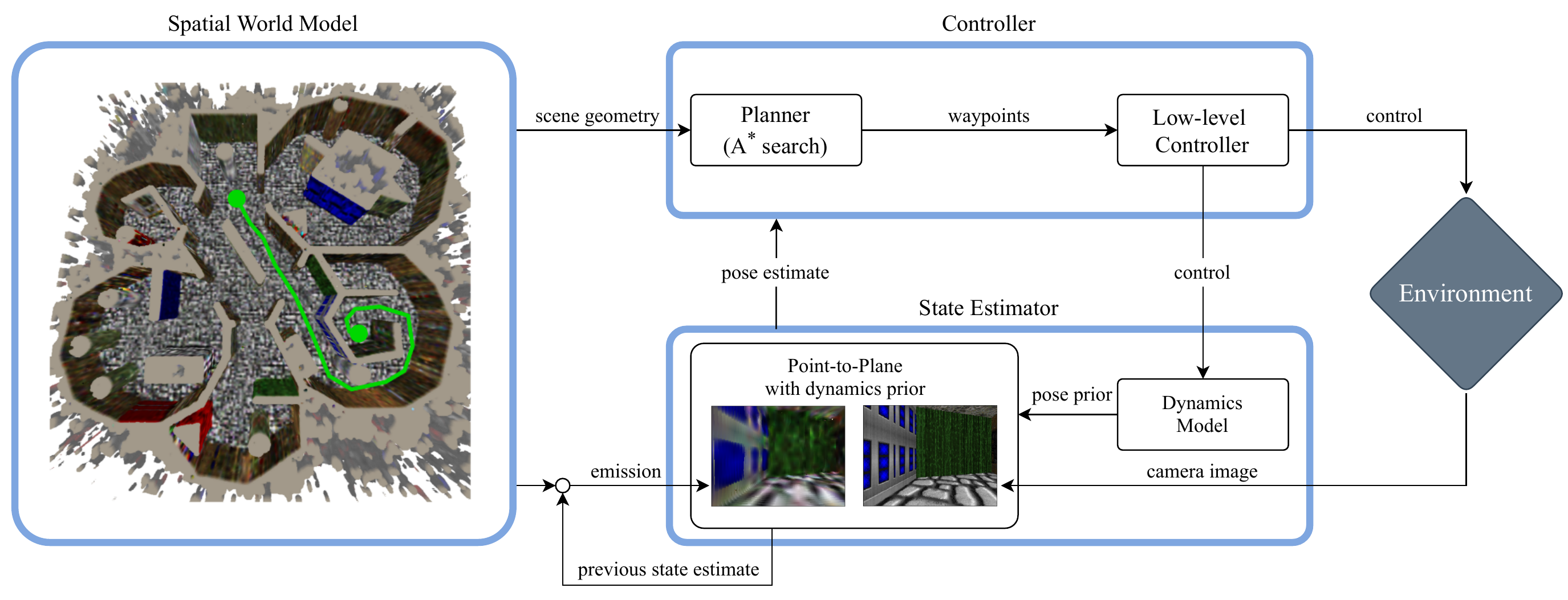}
            \caption{Overview of our method. The planner produces a set of waypoints based on the scene geometry learned by the world model. The low-level controller picks an action to move to the next waypoint. We estimate the agent's state by combining the agent's dynamics with the current camera image, which is aligned against an image predicted by the model.}
            \label{fig:overview}
\end{figure}

We focus on planar indoor robots trying to reach a goal position, the target.
Given a starting location and target coordinates, we plan a trajectory using $\text{A}^*$-search in the space of 2D coordinates (\textit{planning} in \cref{fig:overview}, top).
Here, we discretise the environment using a uniform grid.
Moving from one cell to another is possible only if a) there is no occupied point on the line connecting the two cells according to the occupancy model, and b) the cell we step into maintains a minimum safety distance to any obstacle.
For the former, we sample a fixed (e.g.\ one hundred) number of points along the movement vector.
For the latter, we extract obstacles from the occupancy model by evaluating it on each grid point, where the occupied points then give us a set of point obstacles.
This requires many evaluations of the map, which in our case is computationally cheap due to the voxel grid parameterisation of the map, in contrast to a neural net.
The cost of stepping from one cell to the other is the length of the step and we likewise use the Euclidean distance to the target as our $\text{A}^*$ heuristic.

Our $\text{A}^*$-search yields a shortest path to the target without satisfying 
any constraints on the agent's movement in this phase.
To follow this path, given as a sequence of waypoints, while obeying the limitations of the agent's dynamics, we use a low-level controller (\cref{fig:overview}, top right).
We assume that the agent can rotate in-place up to a maximum angular velocity, and always moves in the direction of its current heading subject to some maximum velocity. 
This allows using a simple low-level controller, which turns the agent at the nearest waypoint and then moves towards it.
Our method can be applied to other systems, as long as a low-level controller capable of following the $\text{A}^*$ plan is available.

\subsection{State Estimation}\label{stateestimation}
Our control rule requires us to continuously track the orientation and 2D location of the agent.
We do so using a tracker that combines a transition model with the RGB-D images observed by the agent (\textit{state estimator} in \cref{fig:overview}).
In the following, we will denote actions with $\control$, see \cref{sec:experiments} for  a description of the assumed noise model for the transition $p(\state_t \mid \state_{t-1}, \control_{t-1})$.
Assuming a point-estimate of $\Map$ we can combine it with the transition into a state-space model:
\eq{
    {p(\state_{2:T}, \obs_{1:T}\mid \control_{1:T-1}, \state_1, \Map)} = {p(\obs_1 \mid \state_1, \Map)}{\prod_{t=2}^T p(\state_{t} \mid \state_{t-1}, \control_{t-1}) p(\obs_t \mid \state_t, \Map)}.
}
We are interested in the filtering posterior $p(\state_t \mid \obs_t, \obs_{1:t-1}, \control_{1:t-1}, \state_1, \Map)$, shortened with $p^t_{\mathrm{filter}}(\state_t)$:
\eq{
    p^t_{\mathrm{filter}}(\state_t) \propto&~ p(\obs_t \mid \state_t, \Map) p(\state_t \mid \obs_{1:t-1}, \control_{1:t-1}, \state_1, \Map) \\
    =&~ p(\obs_t \mid \state_t, \Map)\expcc{p(\state_t \mid \state_{t-1}, \control_{t-1})}{\state_{t-1} \sim\, p^{t-1}_{\mathrm{filter}}(\cdot)}, \numberthis \label{eq:filter}
}
for which we employ a maximum a-posteriori (MAP) approximation:
\eq{
    &\arg\max_{\state_t}\,\, \log p(\obs_t \mid \state_t, \Map) + \log \left( \expcc{p(\state_t \mid \state_{t-1}, \control_{t-1})}{\state_{t-1} \sim\, p^{t-1}_{\mathrm{filter}}(\cdot)} \right) \\
    &\quad \ge \log p(\obs_t \mid \state_t, \Map) + \expcc{\log p(\state_t \mid \state_{t-1}, \control_{t-1})}{\state_{t-1} \sim\, p^{t-1}_{\mathrm{filter}}(\cdot)} \\
    &\quad \approx \log p(\obs_t \mid \state_t, \Map) + \log p(\state_t \mid \state_{t-1}^*, \control_{t-1}). \numberthis \label{eq:seobjective}
}
The second line follows from Jensen's inequality. 
In the last line we approximate the expectation using the previous MAP estimate $\state_{t-1}^*$.

Integrating the agent's desired angular and forward velocities from the action $\control_{t-1}$, we can arrive at a prediction for the current state through $p(\state_t \mid \state_{t-1}^*, \control_{t-1})$, reflected in the second term of \cref{eq:seobjective}.
This prediction is imperfect due to noisy dynamics.
We can refine it by comparing the agent's RGB-D observation against the map by optimising for $p(\obs_t \mid \state_t, \Map)$.
In the literature of differentiable rendering, this is typically done by minimising the difference between the RGB-D observation $\obs_t$ and an image $\hat \obs_t$ rendered from the pose variable $\state_t$ that is subject to optimisation.
This is equivalent to maximising the log-likelihood term $\log p(\obs_t \mid \state_t, \Map)$ above, and we will refer to it as emission-based tracking going forward.

While \cref{eq:seobjective} is well-aligned with the assumed generative model we follow \citep{mirchev2021variational}, it is weighed down by the optimisation of $\log p(\obs_t \mid \state_t, \Map)$ due to the high computational cost of propagating gradients through the renderer.
We provide an empirical analysis of this aspect in \cref{sec:experiments}.
With this in mind, we opt for one more approximation, substituting the first term in \cref{eq:seobjective} for the prediction-to-observation objective introduced in \cref{sec:pointtoplane}:
\eq{
    &\arg\min_{\state_t}\,\, \sum_k \norm{\hat\obs_{t-1}^\rgb[\pi(\transform_{\state_{t}}^{\state_{t-1}^*}\point_{t}^k)] - \obs_{t}^\rgb[\pi(\point_{t}^k)]}_1 \\
    &\quad + \sum_k \norm{\langle \hat\point_{t-1}^k- \transform_{\state_{t}}^{\state_{t-1}^*}\point_{t}^k, \hat\normal_{t-1}^k \rangle}_1 - \log p(\state_{t} \mid \state_{t-1}^*, \control_{t-1}). \numberthis \label{eq:seobjective2}
}
Here, $\transform_{\state_{t}}^{\state_{t-1}^*}$ denotes the relative pose between step $t$ and $t-1$ (going backward in time), which is a function of the optimised $\state_{t}$ and the previous MAP estimate $\state_{t-1}^*$.
With that pose, we reproject points between $\hat\obs_{t-1}^\rgb$, a rendered mean prediction of the RGB image at the previous time step, and the current RGB observation $\obs_{t}^\rgb$.
For the point-to-plane objective, $\point_{t}^k$ is a 3D point in the camera frame of the current observation, computed based on the observed depth $\obs_{t}^\depth$, and $\hat\point_{t-1}^k$ and $\hat\normal_{t-1}^k$ are respectively the corresponding 3D point and normal in the previous camera frame, found through projective data association \citep{blais1995registering}.
The optimised states $\state_{t}$ are parameterised in the $\mathfrak{se}(3)$ Lie-algebra for the special Euclidean group $\mathrm{SE}(3)$, and we assume an L1 norm (Laplace assumption) for robustness to outliers.
We uniformly sample a constant number of pixels for which the objective is evaluated, approximating gradients in expectation.
We perform 100 steps of gradient descent for each new time step using the Adam optimizer \citep{adam}.

By substituting the rendering log-likelihood from \cref{eq:seobjective} for the photometric and point-to-plane loss terms in \cref{eq:seobjective2}, we maintain the necessary geometric constraints in addition to satisfying the assumed transition model.
Note that in the revised objective new observations are still anchored to the map, as the projective transformation happens between a \textit{rendered} prediction $\hat \obs_{t-1}$ and the new incoming observation $\obs_{t}$.
The chosen objective reflects a conditional independence assumption $\obs_{t} \perp\!\!\!\perp \map \mid \obs_{t-1}, \state_{t-1}, \state_{t}$ between the map $\map$ and current observation $\obs_{t}$ given a previous $\obs_{t-1}$ and relative pose offset, which is a reasonable approximation for consecutive time steps.
We opt for this tracking approach in favour of optimising through the emission model as it is faster and just as accurate in our setting, as we will show in experiments.
To the best of our knowledge combining photometric, point-to-plane and dynamics constraints in one objective for state estimation has not been explored before for differentiable rendering.
\section{Experiments} \label{sec:experiments}
\begin{figure}[t]
            \centering
            \includegraphics[width=\linewidth]{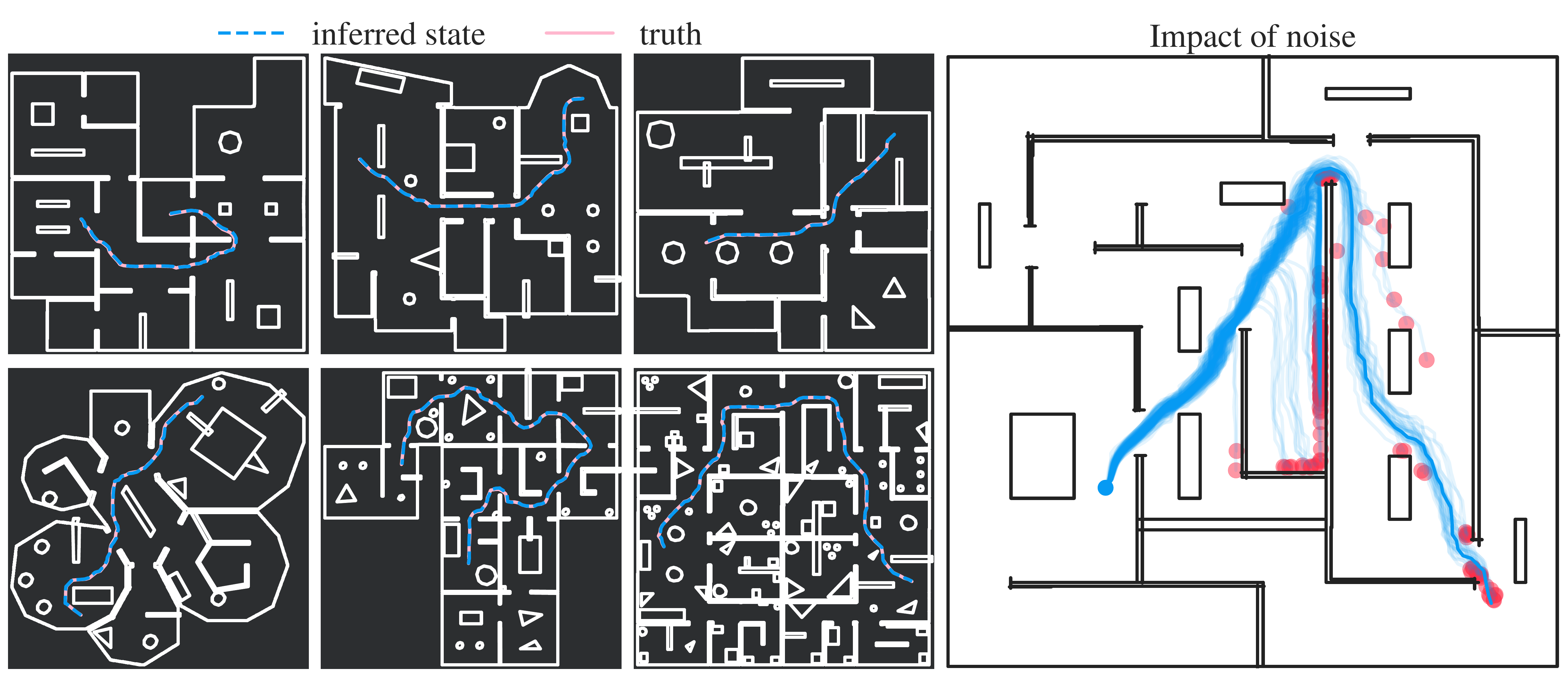}
            \caption{
                Evaluation levels with successful navigation runs (left) and the training maze (right).
                The solid blue line is a reference trajectory. 
                The stochasticity of the dynamics is illustrated by showing multiple samples based on the same control inputs: since each trajectory ends up in a different place, state estimation and closed-loop control are crucial.
            }
            \label{fig:levels}
	    \vspace{-1em}
\end{figure}

We use the Vizdoom simulator for our quantitative experiments \citep{wydmuch2018vizdoom}.
Despite its limited visual fidelity, Vizdoom allows us to focus on environments that are large and complex, i.e. with multiple rooms and realistic floor plans, as we can easily modify the simulation maps.
To that end, we extracted six floor plans from the HouseExpo dataset \cite{li2020houseexpo}, which we converted to Vizdoom levels with added obstacles.
\Cref{fig:levels} shows the evaluation environments, along with a training level taken from \citep{savinov2018semiparametric}.
An agent body length of $\mathrm{0.2m}$, as suggested by \cite{anderson2018evaluation}, corresponds to an environment size of up to a $\mathrm{11.4 \times 11.4 m^2}$ sized square.
A voxel map is learned for each environment from 5000 RGB-D images with pose labels.
The appendix describes the experiment setup and models in further detail.

To analyze navigation performance under stochastic dynamics, we add noise to the agent motion in the simulator.
Here, we make two considerations: noise should only be applied when the agent tries to move and it should not invert the direction of an action.
The latter means that if the agent tries to rotate left, the noise should not make it rotate right.
Likewise, if the agent tries to move forward, noise should not make it move backward.
Guided by these principles, we add clipped Gaussian distributed noise to the angular velocity and speed, where the clipping ensures that the direction of turning (left/right) or movement (forward/backward) is not inverted.

Formally, the controls of the agent are $\mathbf{u} = (\dot{\alpha}, o, s)$, where $\dot{\alpha}$ is the angular velocity, $o$ the angle of the movement direction and $s$ the speed along that direction.
The movement angle is distinct from the current heading angle $\alpha$ but must align with it up to a threshold of $5^\circ$, and there are likewise maximum values defined for the angular velocity and speed at $11.5^\circ$ and $0.8$ times the agent's own body length.
If we take $l$ to be the 2D location of the agent, the noisy dynamics then amount to:
\eq{
    \alpha_{t+1} &= \alpha_t + \text{sign}(\dot{\alpha_t})\max(0, |\dot{\alpha}| + \epsilon_t^\alpha), \hspace{2.5em}\epsilon_t^\alpha \sim \mathcal{N}(0, \sigma_\alpha)}
for the orientation and 
\eq{
    l_{t+1} &= l_t + \begin{bmatrix}\cos (\epsilon^\alpha_t + o_t) \\ \sin(\epsilon^\alpha_t + o_t)\end{bmatrix}\max(0, s_t + \epsilon^s_t), \hspace{0.5em}\epsilon^s_t \sim \mathcal{N}(0, \sigma_s)
}
for the location, where the angular velocity noise is added to the direction of movement and $t$ denotes time.
In addition to the noise, there is another source of error in the agent's dynamics, which results from the simulator.
Vizdoom only allows for setting an integer angle in $[0^\circ, 360^\circ]$ degrees and the 2D location can only be specified up to two decimal points of precision.
Together, these factors make for enough perturbation to pose a non-trivial challenge against navigation.
This effect is illustrated in the right plot of \cref{fig:levels}.
\begin{figure}[t]
    \floatconts{fig:trackingcomp}{
        \caption{
            \vspace{-2em}
            (a,b) CDF of the location \& orientation RMSEs.
            The vertical line shows $0.5 \times \mathrm{agent\_size}$.
            (c) Average runtimes of a single step for different computations: NeRF rendering, voxel map rendering, emission--based tracking, tracking with our scheme, tracking and control.
        }
    }%
    {
        \centering
        \subfigure[][b]{
            \includegraphics[width=0.31\linewidth]{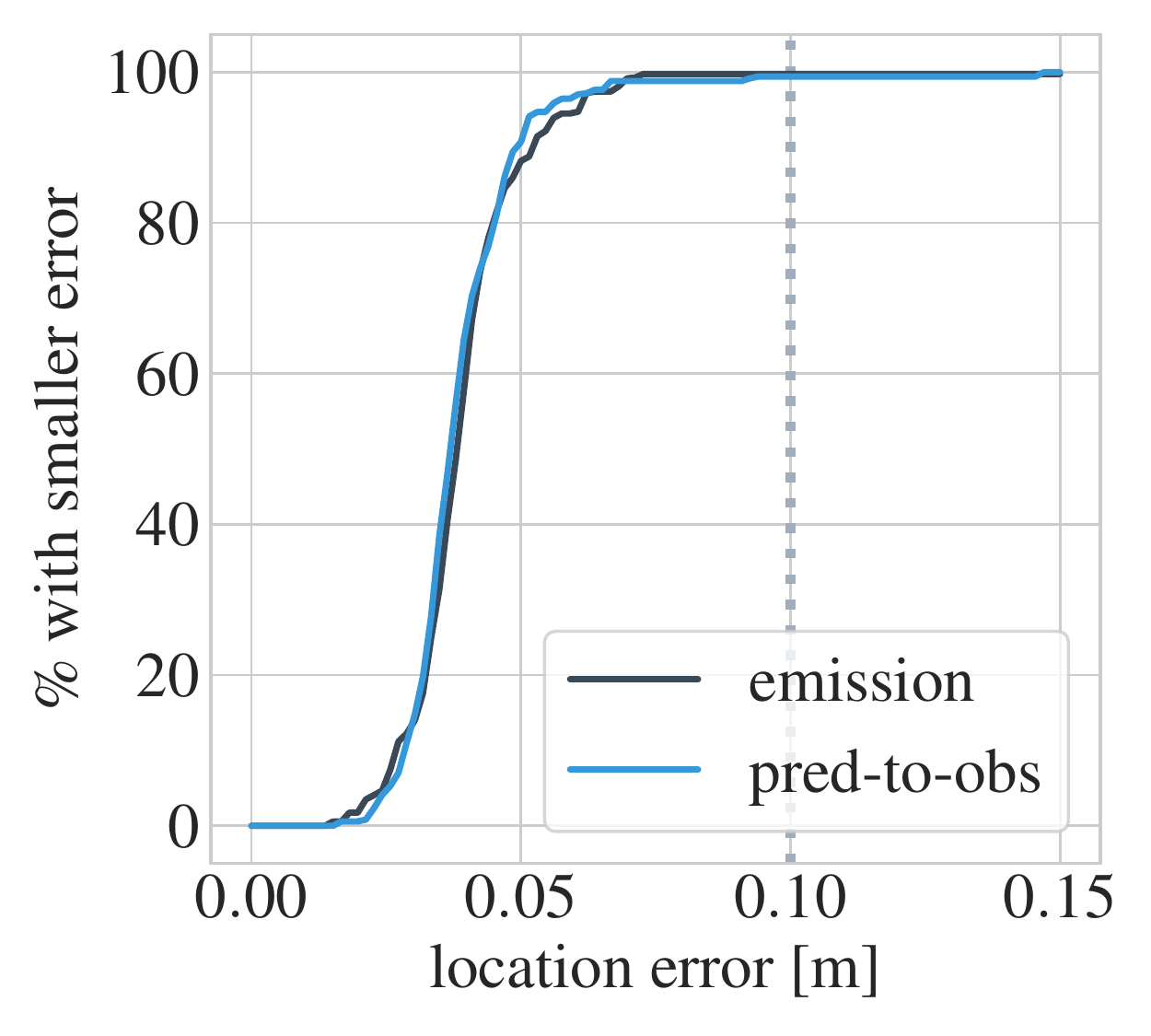}
            \label{fig:trackinglocacc}
        }\hfill
        \subfigure[][b]{
            \label{fig:trackingoriacc}
            \includegraphics[width=0.31\linewidth]{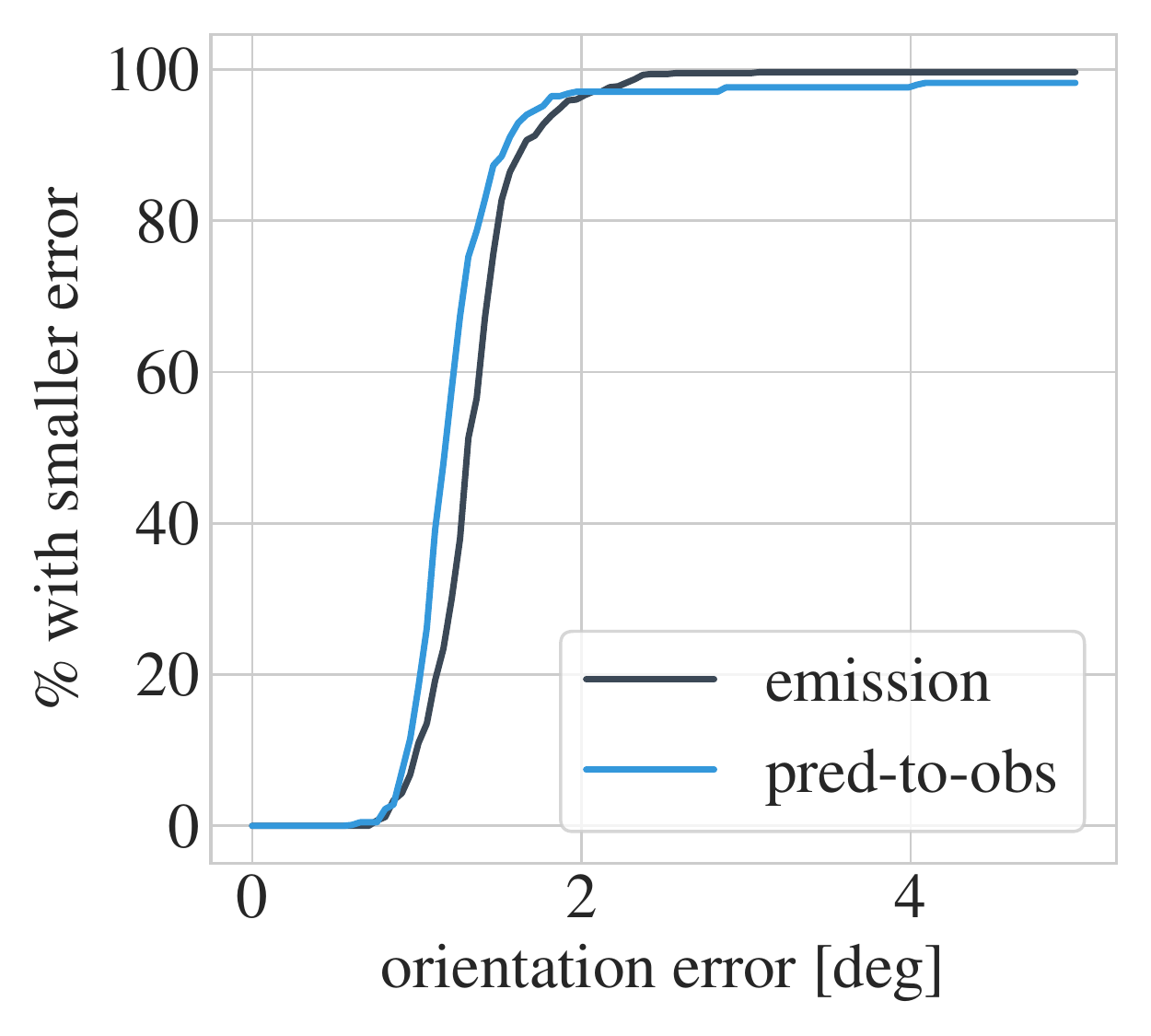}
        }\hfill
        \subfigure[][b]{
            \label{fig:trackingruntimes}
            \includegraphics[width=0.31\linewidth]{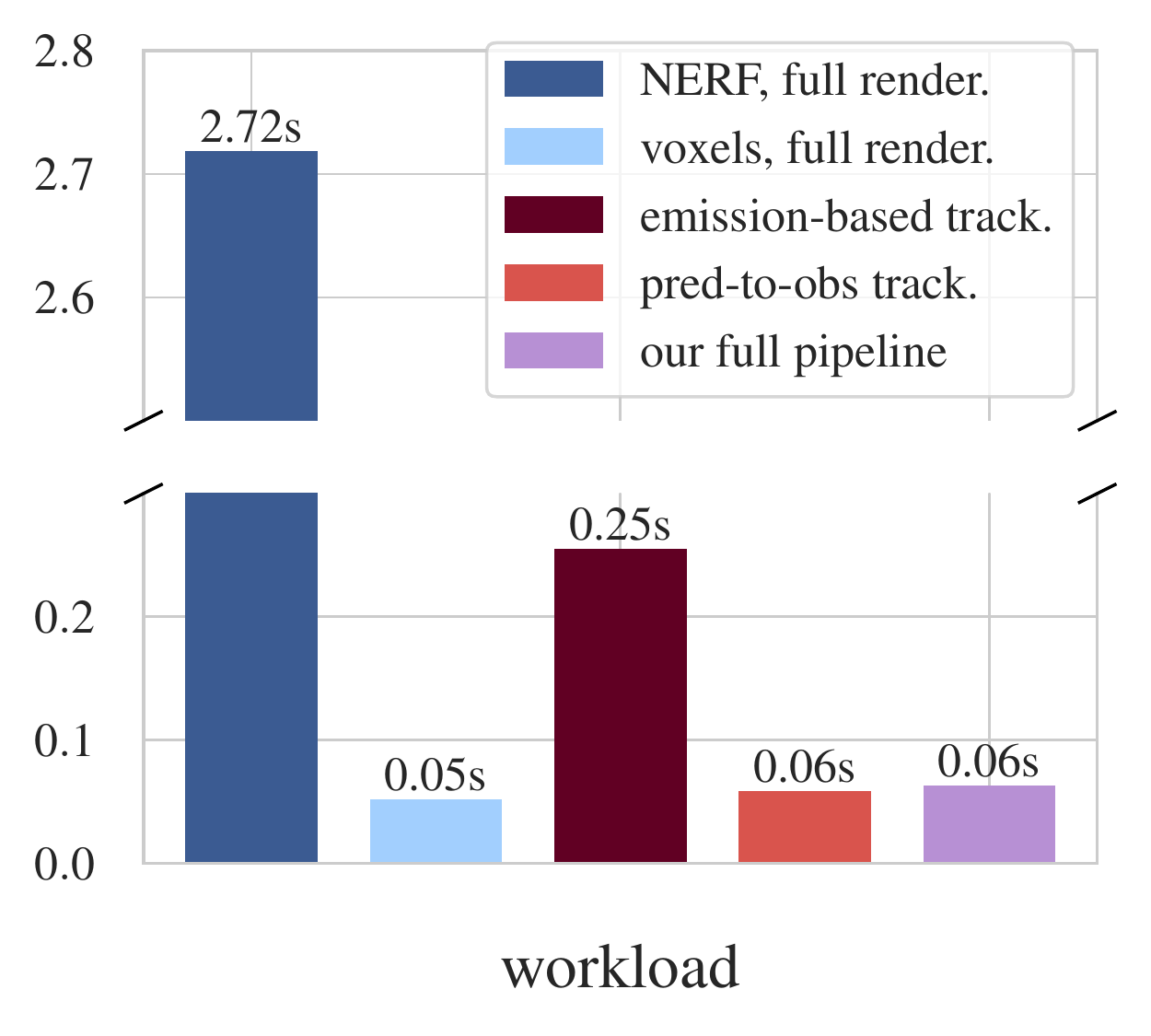}
        }
    }
    \vspace{-1em}
\end{figure}

\paragraph{State estimation in real-time}
First, we compare the tracker described in \cref{stateestimation} to pose estimation via gradient backpropagation through the rendering emission. 
We use $\mathrm{3^\circ}$ and 5\% of the agent body length for the rotational and translational Gaussian scales of the noisy simulator dynamics.
\Cref{fig:trackinglocacc,fig:trackingoriacc} show the accuracy of the two approaches, displaying the percentage of trajectories with an RMSE smaller than the number on the $x$-axis.
Both estimators have adequate performance, with a location RMSE smaller than 0.1m more than 99\% of the time.
Similarly, the orientation RMSEs are roughly on par.

However, the main advantage of the state estimator from \cref{stateestimation} is its speed.
As shown in \cref{fig:trackingruntimes}, a single step takes 0.06s (16.7Hz) on average, compared to 0.25s (4Hz) for the emission-based tracking.
This is due to avoiding the computational cost of propagating gradients through the renderer (cf. \cref{stateestimation}).
We also note the major runtime difference when rendering from a NeRF map \citep{mildenhall2020nerf} at 2.72s (0.4Hz) and from a voxel map \citep{mirchev2021variational} at 0.05s (20Hz) per image (\cref{fig:trackingruntimes}, left) respectively.
The aim for on-line control made us employ the latter approach.
The overall runtime of our pipeline is 0.062s (16.1Hz) per time step, with the bulk of it occupied by the state estimator.
The proposed approach amounts to stable real-time tracking under noisy dynamics with only RGB-D observations.
We believe this is a necessary component in the context of on-line control.

\paragraph{Navigation}
Next we focus on our overarching objective -- successful point-to-point navigation.
Since our tracker works by combining transition estimates with RGB-D observations that are compared against the map, we conduct an ablation study to check how much each part is contributing.
Here, we compare our pipeline to a) tracking by using the transition model only, i.e.\ path integration; b) tracking by optimising a pose offset between two consecutive RGB-D observations (instead of using an emitted image from the map).
\begin{figure}[t]
	\subfigure{
	\begin{tabular}{lccc}
		\toprule
		\textbf{noise} & \textbf{ours} & \textbf{no map} & \textbf {dynamics} \\
        \midrule
		high & $\mathbf{0.46}$ & $0.37$ & $0.33$ \\
		mid & $\mathbf{0.79}$ & $0.51$ & $0.52$ \\
		low & $\mathbf{0.92}$ & $0.61$ & $0.61$ \\
		\bottomrule
\end{tabular}
	\label{table:sr}
}
\subfigure{
	\includegraphics[trim=0 0 0 15cm, clip, scale=0.2, valign=m]{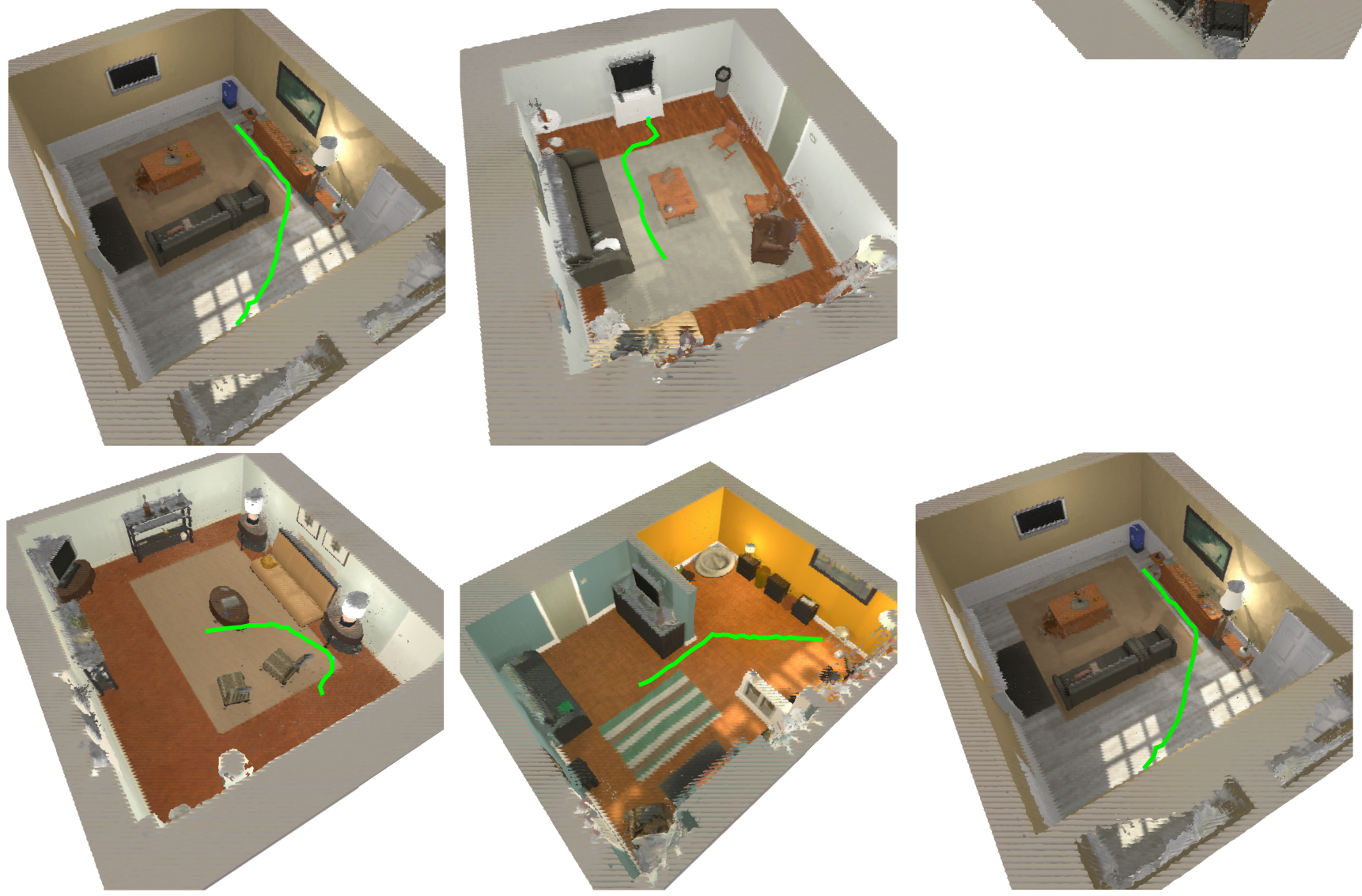}
\label{fig:ai2thor}
}
\caption{
    (a) Average SPL over six environments.
    Higher is better.
    \textbf{No map} tracks by optimising \cref{eq:seobjective2} using the previous camera image instead of the model prediction.
    \textbf{Dynamics} merely integrates the dynamics.
    (b) Isometric views of the learned maps and successful navigation trajectories from AI2-THOR.
    The $\textbf{low}$ noise setting was used.
}
\end{figure}
We repeat our evaluation under three different levels of noise: \textbf{high} ($\sigma_a\myeq9^\circ$, $\sigma_s\myeq30\%$ of agent size), \textbf{medium} ($\sigma_a\myeq6^\circ,\ \sigma_s\myeq15\%$) and \textbf{low} ($\sigma_a\myeq3^\circ,\ \sigma_s\myeq10\%$).

In our evaluation, we sample random pairs of starting positions and targets from the free space of the level, with the constraint that the start and target must be farther apart than three times the agent's body length.
We consider a navigation attempt as successful if the agent's final position is closer to the target than two times the agent's body length, and use 200 navigation tasks per environment.
The same navigation tasks are used for each noise level and method.

Our central evaluation metric is {\it success weighted by path length} (SPL) \citep{anderson2018evaluation}, which is defined as:
\begin{equation}
    \text{SPL} = \sum_{s_i \in \mathcal{S}}s_i \dfrac{l_i}{\max(p_i, l_i)},
\end{equation}
where $\mathcal{S}$ is a set of navigation tasks, $s_i$ is a binary variable indicating the success of a task, and $p_i$ and $l_i$ are the length of the path taken by the agent and the length of the optimal path.
We have found that SPL and success rate are almost the same for all three approaches.
This indicates that successful navigation runs align well with the optimal trajectory.

The results of our evaluation can be found in \cref{table:sr}.
We find that our proposed navigation approach performs best under all noise levels, yet it also degrades at higher levels of noise.
Finally, we qualitatively demonstrate that our work extends to AI2-THOR \citep{ai2thor} in \cref{fig:ai2thor}, 
which is visually much more realistic than Vizdoom.
\section{Conclusion}

We have introduced a method for real-time navigation based on a differentiably rendered world model.
Our approach is able to solve navigation tasks in complex environments using only RGB-D observations and under different levels of noise.
We demonstrated that the state estimation problem can be solved by applying a point-to-plane metric to the differentiable rendering setting and combining it with a dynamics model, which allows for faster tracking compared to differentiating through the renderer.
In future work, we will extend our method to environments with a higher visual fidelity and more complex dynamics.

\bibliography{related}

\clearpage
\appendix{}
\section{Model Details}

We rely on voxel grids to model occupancy and color.
In the Vizdoom experiments, each grid has spatial dimensions of $200\times200\times20$.
For AI2-THOR, we also increase the final dimension to $200$.
The color grid has three feature dimensions that correspond to RGB values.

For Vizdoom we found approximate values for the camera parameters by projecting depth observations and aligning the resulting point cloud with the known level geometry.
Thus, the focal length of the camera is $150$, and the principal offsets in x and y are $160$ and $120$ respectively.
The height and width of a camera image are $320$ and $240$.
The camera is mounted on the body of the player at a quaternion orientation of $[0.5, -0.5, 0.5, -0.5]^T$.
Finally, When casting rays for each pixel, we consider a maximum depth of $20$ and discretise the interval $[0, 20]$ with $200$ points.

In our AI2-THOR experiments, we used a maximum depth of $5$ and picked $50$ equidistant points along each ray.
The camera intrinsic values were taken from the AI2-THOR documentation.
The focal length is set to $\approx 259.8$, corresponding to a field-of-view of $60^\circ$ and an image shape of $300\times300$.
The principal offset in both dimensions is $150$, and the camera-to-body quaternion is $[0.5, 0.5, 0.5, 0.5]^T$.

In both Vizdoom and AI2-THOR we map each environment using a set of $5000$ pose-labeled RGB-D images.
The locations and orientations of these are sampled uniformly from within the free-space of the scene.
For Vizdoom, we optimise the occupancy and color models for $10000$ steps using the Adam optimiser with a learning rate of $0.05$.
These values are reduced to $5000$ and $0.001$ in AI2-THOR.

In either setup, we use a batch size of $25$ and subsample images by picking $200$ pixels.
The scale parameters of the Laplace distribution over color and occupancy $\sigma_1$ and $\sigma_2$ are modeled by setting $\sigma_1 = \sigma_2 / 5$.
With $\sigma_2$ initialised to $2.4$ and thereafter learned using the Adam optimiser with a learning rate of $0.01$.
\section{Motion Planning and Low-level Control Details}

We start $\text{A}^*$-search from the starting location sampled for the respective task and explore the free space of the world model by stepping along the eight cardinal directions.
The step size used here is $80\%$ of the agent's own body length.
We disallow stepping into any location that is closer to any obstacle than a safety distance of $1.2$ times the agent's body length.
The search is stopped as soon as we reach a position that is closer to the goal location than $80\%$ of the agent's body.
Both the occupancy grid we use for checking if a location is navigable and the set of obstacles we use to maintain the safety distance are based on a horizontal slice of the occupancy voxel grid.

The low-level controller first selects the closest waypoint from the trajectory planned by $\text{A}^*$ as a local target.
Given the agent's inferred 2D location $l$, we first define the desired movement vector as $v = l - t$, where $t$ is the local target.
We then check the difference between the angle of this vector and the agent's current heading $o$.
If that difference is below a threshold of $5^\circ$, we allow the agent to take a step forward, respecting a maximum velocity constraint of $80\%$ of the agent's body length.
If the difference is higher than said threshold, the agent stays in-place and rotates to match $o$ with the angle $v$, while obeying a maximum angular velocity constraint.
As soon as the agent is closer than a certain distance to the local target, we remove it from the set of waypoints and pick a new local target.
Finally, if the agent can reach the target by moving in a straight line according to the occupancy model, we define the target as our only landmark, and do no further A*-based planning.
\section{State Estimation Details}\label{app:stateestim}

The agent's starting position is known.
Thereafter, for each control executed in the simulator, we use the imperfect dynamics model to predict the next state of the agent.
Then, we optimise \cref{eq:seobjective2} using this predicted next state, the new camera image received from the simulator and the RGB-D image emitted from the model using the last inferred state.
We use the Adam optimiser for $100$ steps with a learning rate of $0.02$.
At every optimization step, we compute the loss for $1000$ randomly selected pixels from the new image observation $\obs_t$.
We sample a different set of pixels in each of the $100$ optimisation steps, but keep these $100\times1000$ pixel indices fixed for all invocations of the state estimator.
Note that the subsampling happens only in the observed image, the predicted previous-step image $\hat\obs_{t-1}$ needs to remain whole, as we need its gradient information.
We use bilinear interpolation of $\hat\obs_{t-1}$ to make the photometric term in \cref{eq:seobjective2} differentiable w.r.t.\, the camera pose.
We use the numeric gradients of $\hat\obs_{t-1}$ to compute the normals necessary for the second point-to-plane term in \cref{eq:seobjective2}.
For the sake of robustness, we threshold the absolute values of the geometric and photometric errors in \cref{eq:seobjective2}.
For a pixel with index $k$, the photometric error term was defined in \cref{eq:seobjective2} as:
\eq{
	\norm{\hat\obs_{t-1}^\rgb[\pi(\transform_{\state_{t}}^{\state_{t-1}^*}\point_{t}^k)] - \obs_{t}^\rgb[\pi(\point_{t}^k)]}_1.
}
We ignore this term for any pixel where it exceeds a value of $1.0$.
Likewise, the geometric term is:
\eq{
	\norm{\langle \hat\point_{t-1}^k- \transform_{\state_{t}}^{\state_{t-1}^*}\point_{t}^k, \hat\normal_{t-1}^k \rangle}_1,
}
and it too is ignored for any pixel where it is higher than $5.0$.
\section{Navigation Experiments}

We sample $200$ navigation tasks from each environment.
Here, we only allow tasks where the start and target are farther apart than three times the agent's body length.
Likewise, we do not allow start and target locations that are closer to any obstacle than $1.2$ times the agent's body.
For this part, we access the actual scene geometry of the simulator in Vizdoom to define obstacles.
Since this is not possible in AI2-THOR, we instead use the learned occupancy map.

In Vizdoom, we implement continuous control via the teleportation and rotation commands.
These do not check for collisions, so we check these manually by calculating distances to the obstacles, which we extract from the simulator in the form of a set of line segments.
As soon as a collision occurs, we prohibit further movement.
AI2-THOR likewise provides a teleportation command which can also be used to set rotations but unlike Vizdoom it also checks for collisions.
Thus, in AI2-THOR we use the simulator's own collision handling.

The three images shown in \cref{fig:ai2thor} are based on the following AI2-THOR scenes: Floorplan215, FloorPlan224, FloorPlan209.
\section{Tracking Experiments}

We evaluate the two considered state estimators using ground-truth pose data for 170 successful navigation trajectories in our training Vizdoom environment (\cref{fig:levels}, right).

First, we tune both the emission-based state estimator and the prediction-to-observation state estimator from \cref{stateestimation}, searching for their optimal optimisation step sizes.
We allow a different step size for the translation and rotation component of the optimised camera pose.
We perform a hyperparameter search with 4000 runs for each tracker, optimising for location and orientation RMSE for the first 10 trajectories.
We found the optimal step size for the emission-based state estimator to be $(0.001, 0.002)$ for translation and rotation optimisation, respectively.
We found the optimal step size for the prediction-to-observation state estimator to be $(0.01, 0.01)$ for translation and rotation optimisation, respectively.

After tuning, we evaluate both estimators on all data, running for 5 random seeds for each trajectory to account for stochasticity, and then we aggregate the results.
We report absolute location and orientation RMSE errors computed between the predicted tracked trajectories and the ground-truth trajectory poses.
All other hyperparameters remain as described in \cref{app:stateestim}.

When measuring runtimes, we isolate an NVIDIA Quadro RTX 8000 GPU and 8 Intel Xeon Gold 6252 CPU cores at 2.1GHz in a cluster environment.
We then run each state estimator on 10 trajectories, measuring the runtime of each tracking step, and report the respective aggregated results.

\end{document}